\documentclass{article}

\usepackage{arxiv}

\usepackage[utf8]{inputenc} 
\usepackage[T1]{fontenc}    
\usepackage{hyperref}       
\usepackage{url}            
\usepackage{booktabs}       
\usepackage{amsfonts}       
\usepackage{nicefrac}       
\usepackage{microtype}      
\usepackage{lipsum}		
\usepackage{graphicx}
\usepackage{natbib}
\usepackage{doi}
\usepackage{alltt}

\usepackage{xcolor}

\title{Annotation guidelines for the Turku Paraphrase Corpus}

\author{Jenna Kanerva, Filip Ginter, Li-Hsin Chang, Iiro Rastas, Valtteri Skantsi, Jemina Kilpel{\"a}inen,\\ \textbf{Hanna-Mari Kupari, Aurora Piirto, Jenna Saarni, Maija Sev{\'o}n, and Otto Tarkka} \\
  TurkuNLP Group \\
  Department of Computing \\
  Faculty of Technology \\
  University of Turku, Finland \\
  {\tt jmnybl@utu.fi}}

\date{}

\renewcommand{\headeright}{Technical Report}
\renewcommand{\undertitle}{Technical Report}
\renewcommand{\shorttitle}{}

\begin{document}
\maketitle

\begin{abstract}
	This document describes the annotation guidelines used to construct the Turku Paraphrase Corpus. These guidelines were developed together with the corpus annotation, revising and extending the guidelines regularly during the annotation work. Our paraphrase annotation scheme uses the base scale 1--4, where labels 1 and 2 are used for negative candidates (not paraphrases), while labels 3 and 4 are paraphrases at least in the given context if not everywhere. In addition to base labeling, the scheme is enriched with additional subcategories (flags) for categorizing different types of paraphrases inside the two positive labels, making the annotation scheme suitable for more fine-grained paraphrase categorization. The annotation scheme is used to annotate over 100,000 Finnish paraphrase pairs.
\end{abstract}


\section{Introduction}

This document describes the annotation guidelines used to construct the Turku Paraphrase Corpus\footnote{The Turku Paraphrase Corpus is available at \url{https://turkunlp.org/paraphrase.html}.}, a corpus of Finnish paraphrases including over 100,000 manually annotated paraphrase pairs. The first release of the corpus is described in \citet{kanerva2021paraphrase}. These annotation guidelines were developed simultaneously with the corpus annotation, together with all annotators involved in the corpus construction. Guidelines were revised and extended regularly during the annotation work.

The paraphrase annotation scheme uses the base scale 1--4, where labels 1 and 2 are used for negative candidates (not paraphrases), while labels 3 and 4 are paraphrases at least in the given context if not everywhere. In addition to base labeling, scheme is enriched with additional subcategories (flags) for categorizing different types of paraphrases inside the two positive labels. These flags account for e.g.\ one statement being slightly more detailed than the other, or one being ambiguous in meaning, statements having tone or register differences, or other minor deviations in meaning.

The guidelines are organized as follows. First we give short definitions of the base labels and flags used in the annotation, as well as list few principal guidelines used throughout the annotation work. Then we continue to explain each label in a detailed fashion including example annotations in Finnish as well as their English translations\footnote{Note that the provided translations are only reference translations not necessarily fully reflecting phenomena demonstrated in the Finnish examples.}. Finally, we explain additional annotation guidelines outside the annotation scheme relevant for the annotation workflow of the Turku Paraphrase Corpus.

\section{Label definitions}

\paragraph{Label 4:} Universal (perfect) paraphrase in all reasonably imaginable contexts, meaning one can always be replaced with the other without changing the meaning. This ability to substitute one for the other in any context is the primary test for label 4 used in the annotation.
\paragraph{Label 3:} Context dependent paraphrase, where the meaning of the two statements is the same in the present context, but not necessarily in other contexts.
\paragraph{Label 2:} Related but not a paraphrase, where there is a clear relation between the two statements, yet they cannot be considered paraphrases.
\paragraph{Label 1:} Unrelated, there being no reasonable relation between the two statements, most likely a false positive in candidate selection.
\paragraph{Label x:} Skip if labeling is not possible for a reason, or giving a label would not serve the desired purpose (e.g. wrong language or identical statements).

With these base labels alone, a great number of candidate paraphrases would fail the substitution test for \emph{label 4} and would be classified with \emph{label 3}. In order to avoid populating the \emph{label 3} category with a very diverse set of paraphrases, we opt to introduce flags for finer sub-categorization. The additional flags can be only attached to \emph{label 4} (subcategories of universal paraphrases), meaning the paraphrases are not fully interchangeable due to the specified reason, but, crucially, are context-independent, unlike \emph{label 3}. The possible flags are:

\paragraph{Subsumption (\textgreater~or \textless):} One of the statements is more detailed and the other more general. The relation of the pair is therefore directional, where the more detailed statement can be replaced with the more general one in all contexts, but not the other way around.
\paragraph{Style (s):} Tone or register difference in cases where the meaning of the two statements is the same, but the statements differ in tone or register such that in certain situations, they would not be interchangeable.
\paragraph{Minor deviation (i):} Minimal differences in meaning (typically ”this” vs. ”that” ) as well as easily traceable differences in grammatical number, person, tense or such. The desired treatment of these examples is considered highly dependent on the downstream application.

The flags are independent of each other, thus each example can be annotated with a combination of different flags. The exception to this rule is the directional subsumption, where a justification for crossing directionality (one statement being more detailed in one aspect while the other in another aspect) decreases the base label into \emph{label 3} as the directional replacement test does not hold anymore.

\section{General instructions}
\begin{itemize}

    \item Focus on semantics instead of syntax.
    \item Typo level differences are ignored if the mistake in spelling does not create ambiguities in meaning.
    \item Flags are subcategories of base label 4 and flags cannot be attached to context dependent paraphrases (\emph{label 3}). If a flag (or combination of flags) is enough for accounting the difference(s), prefer \emph{label 4 + flag} over \emph{label 3}. 
    \item Prefer a higher label in cases where two labels seem well justified. For example in cases, where you would not use a word as a synonym yourself but can imagine someone else using, treat the words in question as synonyms.

\end{itemize}

\section{Label 4: Universal paraphrases}

\begin{itemize}
    
\item The utterances are paraphrases in all non-absurd situations.
\item Utterances can be alternated without changing the meaning in all contexts.

    \begin{alltt}
        Tulen puolessa tunnissa
        {\footnotesize'I'll be there in half an hour'}
        Saavun 30 minuutin kuluessa 
        {\footnotesize'I will arrive in 30 minutes'} →4

        Voitinko minä?
        {\footnotesize'Did I win?'}
        Olenko voittaja?
        {\footnotesize'Am I the winner?'} →4 

        Tykkään kasvojesi painaumista.
        {\footnotesize'I like the dents on your face.'}
        Pidän noista kuopista kasvoissasi 
        {\footnotesize'I enjoy those impressions on your face.'} →4
    \end{alltt}

\item Merely syntactic difference is not accounted, e.g. lack of discourse connective.

    \begin{alltt}
        Mutta kolme kertaa? Silloin karma yrittää kertoa jotakin.
        {\footnotesize'But three times? Then karma is trying to tell you something.'}
        Kolme kertaa tarkoittaa, että karma yrittää kertoa jotain. 
        {\footnotesize'Three times means that karma is trying to tell you something.'} →4 
    \end{alltt}
    
\item Absurd interpretations of an utterance are not considered. Also idioms and phrasal expressions can have a different literal meaning, but if the literal meaning is absurd it is not taken into account. Also clear metaphors and simile are \emph{label 4} when the literary meaning is absurd. Usually the word “kuin” indicates a simile.

    \begin{alltt}
        Tämä aika vuodesta on pahasta hiuksilleni.
        {\footnotesize'This time of year is bad for my hair.'}
        Hiukseni eivät pidä tästä vuodenajasta.    
        {\footnotesize'My hair does not like this time of year.'} →4

        voin heittää sinut kotiin
        {\footnotesize'I can give you a ride home'}
        pääset minun kyydissäni kotiin 
        {\footnotesize'you can catch a lift home with me'} →4

        Työskentelee kuin ahkera mehiläinen
        {\footnotesize'He/she is working busy as a bee'} 
        Ahertaa kuin muurahainen            
        {\footnotesize'Toiling away like an ant'} →4
    \end{alltt}

\item The difference in number, tense or similar is not taken into consideration in metaphors or simile since it is part of the expression and does not affect meaning, same for plurale tantum expressions.

    \begin{alltt}
        Se matkaa tähtien välillä kuin aurinkotuulet
        {\footnotesize'It is traveling between the stars like a solar wind'}
        Se leviää tähtien välissä kuin kulovalkea   
        {\footnotesize'It is catching on among the stars like wildfire'} →4 
        
        Häät
        {\footnotesize'Wedding'}
        Hääjuhla 
        {\footnotesize'Wedding party'} →4

        Olen vain idiootti, joka höpöttää puhelinvastaajaan
        {\footnotesize'I'm just a fool who is blabbing on an answering machine'}
        Olen niitä idiootteja jotka höpöttävät vastaajaan  
        {\footnotesize'I'm one of those idiots who blab on answering machines'} →4 
        
        Sain ajatuksen.
        {\footnotesize'I have got an idea'}
        Minulla on idea.
        {\footnotesize'I have a thought.'} →4

        \emph{Compare to}
        Parkkeerasimme tähän
        {\footnotesize'We parked here'}
        Parkkeeraamme tähän 
        {\footnotesize'We will be parking here'} →4i
    \end{alltt}

\item Proper nouns can be considered the same as their common noun descriptions when there is no other interpretation based on world knowledge, also specific events or names that are general knowledge and easily findable can be considered as paraphrases. However, if the event is not individualized into a single occasion and it has or could happen again, it is not considered universal paraphrase in all situations.

    \begin{alltt}
        Ensimmäinen avaruuteen lähetetty suomalaissatelliitti tuhoutui
        {\footnotesize'The first Finnish satellite sent into space was destroyed'}
        Aalto-2 tuhoutui                                               
        {\footnotesize'Aalto-2 was destroyed'} →4
        
        \emph{Compare to}
        Turun arvokuljetusryöstö
        {\footnotesize'The Turku robbery'}
        Vuoden 2002 Turun arvokuljetusryöstö 
        {\footnotesize'The 2002 robbery in Turku'} →4<
    \end{alltt}
    
\item Pronoun references and noun descriptions can be considered the same in cases where there is no clear difference in specificity.

    \begin{alltt}
        Tyyppi tappoi Fisherin.
        {\footnotesize'This person killed Fisher.'}
        Hän tappoi Fisherin.    
        {\footnotesize'He/she killed Fisher.'} →4

        \emph{Compare to}
        Tämä tyyppi tappoi Fisherin.
        {\footnotesize'This person here killed Fisher.'}
        Hän tappoi Fisherin.         
        {\footnotesize'He/she killed Fisher.'} →4>
    \end{alltt}

\item If there is a pronoun or word that holds no or only very little semantic meaning in that context, it can be ignored in evaluation. Also minor differences are not taken into account, e.g. missing personal pronoun if the person is detectable from other parts of the utterance or its meaning.

    \begin{alltt}
        Mutta huomaan kyllä, ettei tämä ole ensimmäinen kerta, kun suusi on
        johtanut sinut ongelmiin.
        {\footnotesize'I realize that this is not the first time when your mouth has lead you to
          trouble.'} 
        Mutta, näen ettei ole ensimmäinen kerta kun suusi on saanut sinut
        vaikeuksiin.                                                        
        {\footnotesize'I can see that it's not the first time your mouth has caused you difficulties.'}→4
        
        Loukkuun jäätiin.
        {\footnotesize'Trapped in here.'}
        Olemme ansassa.  
        {\footnotesize'We are in a snare.'} →4
    \end{alltt}

\item Sarcastic expressions are considered to be universal paraphrases when it is clear from the context or expression that they can not be interpreted for their literal meaning.

    \begin{alltt}
        Etsin "ystävääni", joka ampui minua.
        {\footnotesize'I'm looking for the "friend" who shot me.'}
        Etsin niin sanottua ystävääni. Hän ampui minua. 
        {\footnotesize'I'm looking for my so called friend. He/she shot me.'} →4
    \end{alltt}
    
\item Euphemistic expressions are treated as a paraphrase when their semantic content is clear and unambiguous. 

    \begin{alltt}
        Kai sinä pystyt sänkyhommiin?
        {\footnotesize'I hope you can perform in bed?'}
        Olethan täysin kykenevä seksiin? 
        {\footnotesize'I hope that you are capable of having sex?'} →4
    \end{alltt}
    
\item Label 4 can be assigned also if both utterances have a similar permanent title (e.g. honorary title) or the permanent title can be considered world knowledge. However, in case of temporary titles (e.g. prime minister), the title is considered as additional information.

    \begin{alltt}
        Sussexin herttua Harry
        {\footnotesize'Harry, Duke of Sussex'}
        Britannian Harry      
        {\footnotesize'Harry of Britain'} →4
        
        Jalkapalloilija Diego Maradona
        {\footnotesize'Football player Diego Maradona'}
        Diego Maradona →4
        

        \emph{Compare to} 
        Presidentti Niinistö matkusti
        {\footnotesize'President Niinistö traveled'}
        Niinistö matkusti             
        {\footnotesize'Niinistö traveled'} →4>
    \end{alltt}

\item If one of the utterances is composed of two separate elements (e.g. two sentences), the actor of the sentences can be assumed to be the same person in both instances if not clearly otherwise stated. This holds also in cases where the actor would be ambiguous (when occurring alone), if the reference is clear from the additional part.

    \begin{alltt}
        Niinistölle poikavauva – Presidenttipari onnessaan
        {\footnotesize'The Niinistö family has a boy - Presidential couple is happy'}
        Presidentti Sauli Niinistö puolisoineen onnellinen poikavauvan johdosta →4
        {\footnotesize'The president Sauli Niinistö and his spouse are happy after having a baby boy'}

        \emph{Compare to}
        Niinistölle poikavauva
        {\footnotesize'Niinistö has a baby boy'}
        Presidentti Sauli Niinistölle poikavauva 
        {\footnotesize'President Sauli Niinistö has a baby boy'} →4<
    \end{alltt}

\item Label 4 can be assigned if both utterances have the same register, stylistic features, and semantic meaning --- even if it is vulgar, and even if there is a small difference in strength while both still being vulgar. However, if the target of the vulgar expression is different (person or thing), it may cause the subsumption flag or label 3.

    \begin{alltt}
        Hiljaa, paskiaiset!
        {\footnotesize'Quiet, you bastards!'}
        Turpa kiinni, kusipäät! 
        {\footnotesize'Shut up you assholes!'} →4

        Vitustako minä tiedän.
        {\footnotesize'How the fuck should I know.'}
        Mistä pirusta minä tiedän. 
        {\footnotesize'How the hell should I know.'} →4

        \emph{Compare to}
        Hitot.
        {\footnotesize'Damn it.'}
        Haista vittu. 
        {\footnotesize'Fuck you.'} →3
    \end{alltt}

\item Personal evaluation vs evaluative statement; in some contexts, an evaluative statement can be read as a personal evaluation even when a personal pronoun is not explicitly used, in these cases the label 4 can be assigned instead of the subsumption flag.

    \begin{alltt}
        pidän mekostasi
        {\footnotesize'I like your dress'}
        sinulla on hieno mekko 
        {\footnotesize'you have a fine dress'} →4

        Kiva, että olet saanut töitä
        {\footnotesize'It's nice that you have found a job'}
        Olen iloinen, että sait työn 
        {\footnotesize'I'm happy that you got the job'} →4

        Ketään ei tunnu olevan kotona.
        {\footnotesize'Nobody seems to be at home.'}
        En usko kenenkään olevan kotona. 
        {\footnotesize'I don't believe that there is anyone at home.'} →4

        \emph{Compare to}
        Minusta hän ei ole tarpeeksi hyvä sinulle
        {\footnotesize'I don't think that he is good enough for you'}
        Hän ei ole kyllin hyvä sinulle            
        {\footnotesize'He is not adequate for you'} →4> 

        \emph{Compare to}
        Taidamme päästä kärryille kieliopin suhteen.
        {\footnotesize'We are getting a hang of the grammar.'}
        Luulisin, että alamme ymmärtään kielioppia. 
        {\footnotesize'I believe that we are starting to understand grammar.'} →4<
    \end{alltt}

\item Label 4 can be assigned if there is no clear difference in style or tone or specificity. Some words hold only little semantic meaning that makes them redundant.

    \begin{alltt}
        Kiitti, se oli tarpeen
        {\footnotesize'Thanks, it was necessary'} 
        Kiitos tarvitsin sitä 
        {\footnotesize'Thank you, I needed that'} →4

        Älä koskaan kyseenalaista käskyjäni muiden edessä.
        {\footnotesize'Never undermine my orders in front of the others.'}
        Älä kyseenalaista käskyjäni muiden edessä.        
        {\footnotesize'Don't undermine my orders in front of the others.'} →4

        Halusin sinun kuulevan sen minun suustani.
        {\footnotesize'I wanted for you to hear it from me.'}
        Halusin, että kuulet asian minulta.       
        {\footnotesize'I wanted for you to get the information from me.'} →4

        Onko se totta?
        {\footnotesize'Is that true?'}
        Onko asia niin? 
        {\footnotesize'Are the things that way?'} →4
    \end{alltt}

\end{itemize}

\section{Subsumption flag (\textgreater~or \textless)}
\begin{itemize}

\item The subsumption flag is assigned when one utterance is more general than the other.

\item More general utterance means that the utterance can be implied from the other utterance, i.e. it conveys less information and can be interpreted and used in more ways than the utterance in comparison. The most common reasons for the utterance being more general is it missing some minor additional information the other utterance includes, or it being ambiguous in a way the other utterance is not. Here the utterances are interchangeable in one direction only. A justification for crossing directionality (one utterance being more detailed in one aspect while the other in another aspect) decreases the base label into \emph{label 3} and the subsumption flag is not used.

\item The $>$ symbol means that the latter utterance is more general, i.e. the arrow points to the more general utterance.
\item The $<$ symbol means that the former utterance is more general.

    \begin{alltt}
        Haluan vain puhua.
        {\footnotesize'I just want to talk.'}
        Tahdon vain puhua hänen kanssaan. 
        {\footnotesize'I just want to talk to him/her.'} →4<

        En tiedä, haluanko naimisiin Hamishin kanssa.
        {\footnotesize'I don't know if I want to marry Hamish.'}
        En tiedä, haluanko naimisiin.                
        {\footnotesize'I don't know if I want to get married.'} →4>

        Ei minua nukuta
        {\footnotesize'I'm not sleepy.'}
        En ole väsynyt 
        {\footnotesize'I'm not tired.'} →4>

        Jatka ajamista.
        {\footnotesize'Keep on driving.'}
        Jatka matkaa.  
        {\footnotesize'Keep on going.'} →4>

        Posti lisää pakettiautomaatteja ja karsii myymälöitään
        {\footnotesize'The Posti Group is increasing the amount of parcel lockers and diminishing the
          number of its shops'} 
        Turbulenttinen Posti lisää pakettiautomaatteja ja karsii myymälöitään. 
        {\footnotesize'The turbulent Posti Group is increasing the amount of parcel lockers and
          diminishing the number of its shops'} →4<
    \end{alltt}
    
\item The other utterance has a pronoun or possessive suffix which the other utterance is lacking, and which cannot be derived from other words or is not otherwise clear from the context.

    \begin{alltt}
        Sinun täytyy antaa luottokorttisi numero, mutta tämä on hauskaa.
        {\footnotesize'You have to give your credit card number but this is fun.'}
        No, sinun täytyy laittaa luottokortin numero, mutta tämä on hauskaa. 
        {\footnotesize'You have to give a credit card number but this is fun.'} →4>

        Loistava ajoitus 
        {\footnotesize'Perfect timing'}
        Tulit juuri sopivasti 
        {\footnotesize'You arrived just perfectly'} →4<

        \emph{Compare to}
        Neljävuotias poika jäi isovanhemman kuljettaman traktorin alle 
        {\footnotesize'A four year old boy was trapped under a tractor driven by a grandparent'}
        4-vuotias poika loukkaantui jäätyään isovanhempansa kuljettaman
        traktorin alle                                                  
        {\footnotesize'A four year old boy trapped under a tractor driven by one of his grandparents'} →4
    \end{alltt}

\item The other utterance convey a verb with a person agreement, which the other is lacking (typically passive form), and the person cannot be derived from the meaning.

    \begin{alltt}
        Olemmeko aikataulussa?
        {\footnotesize'Are we on the set time schedule?'} 
        Ollaanko aikataulussa? 
        {\footnotesize'Is the time schedule holding?'} →4> 

        \emph{Compare to}
        Mitä teette täällä? 
        {\footnotesize'What are you doing here?'}
        Mitä te täällä teette? 
        {\footnotesize'What are you guys doing here?'} →4

        \emph{Compare to}
        Loukkuun jäätiin.
        {\footnotesize'Trapped in here.'}
        Olemme ansassa.  
        {\footnotesize'We are in a snare.'} →4
    \end{alltt}

\item Also second person passive vs regular passive is marked with the subsumption flag as the second person passive can be considered ambiguous in meaning (passive or genuine second person).

    \begin{alltt}
        Joskus amerikkalaisuus tarkoittaa, että pyydät anteeksi. 
        {\footnotesize'Sometimes being an American means that you say you are sorry.'} 
        Toisinaan amerikkalaisuus tarkoittaa anteeksipyyntöä.    
        {\footnotesize'There are times when being an American means apologizing.'} →4>
    \end{alltt}

\item The other utterance has a pronoun, whereas the other has a noun or proper noun and there is clear difference in specificity.

    \begin{alltt}
        Hänestä on tullut oikea pulisija.
        {\footnotesize'He has started talking non-stop.'}
        Pekasta on tullut melkoinen papupata. 
        {\footnotesize'Pekka is so talkative these days.'} →4<

        Compare to
        Onko se totta?
        {\footnotesize'Is that true?'}
        Onko asia niin? 
        {\footnotesize'Are the things that way?'} →4
    \end{alltt}

\item The other utterance has an ambiguous, context dependent adverb, while the other is more specific.  

    \begin{alltt}
        Olemme täällä
        {\footnotesize'We are here'}
        Olemme ulkona 
        {\footnotesize'We are outside'} →4<

        \emph{Compare to}
        Olemme täällä
        {\footnotesize'We are here'}
        Olemme tuolla 
        {\footnotesize'We are over there'} →4i
    \end{alltt}

\item Sarcastic expressions that can be interpreted in two ways (sarcastic and literal meaning) are given a subsumption flag if the paraphrase pair is not ambiguous in the same way. However, if the utterance does not have any visible signal of possible sarcasm, it is not interpreted as being such.

    \begin{alltt}
        Joopa joo, olet täysin puolueeton Lanan suhteen.
        {\footnotesize'Oh, yeah right, you are totally impartial when it comes to Lana.'}
        Olet täysin objektiivinen Lanaa koskevissa asioissa. 
        {\footnotesize'You are completely objective when it comes to Lana.'} →4<

        \emph{Compare to}
        olet täysin puolueeton Lanan suhteen.
        {\footnotesize'You are totally impartial when it comes to Lana.'}
        Olet täysin objektiivinen Lanaa koskevissa asioissa. 
        {\footnotesize'You are completely objective when it comes to Lana.'} →4
        
        \emph{Compare to}
        Etsin "ystävääni", joka ampui minua.
        {\footnotesize'I'm looking for the "friend" who shot me.'}
        Etsin niin sanottua ystävääni. Hän ampui minua. 
        {\footnotesize'I'm looking for my so called friend. He/she shot me.'} →4
    \end{alltt}

\item The other utterance is a genuine open question, while the other makes an assumption of what the answer is likely to be or is a rhetorical question.

    \begin{alltt}
        Seuraatko urheilua?
        {\footnotesize'Do you follow sports?'}
        Taidat olla urheilufani. 
        {\footnotesize'You seem like a sports fan, don't you?'} →4<

        Etkö sinä nähnyt sitä?
        {\footnotesize'You didn't see that?'}
        Et nähnyt sitä, ethän?  
        {\footnotesize'I hope you did not see that?'} →4<

        Ethän kerro kenellekään?
        {\footnotesize'I hope you are not telling anyone?'}
        Älä kerro kenellekään. 
        {\footnotesize'Don't tell anyone'} →4<

        \emph{Compare to}
        Lupaathan, ettei tämä toistu?
        {\footnotesize'You will promise me that this will not be repeated?'}
        Lupaa, ettei tämä toistu. 
        {\footnotesize'Promise me that this will not be repeated?'} →4s

        \emph{Compare to}
        Tämä ei ole ensimmäinen kerta, kun suusi on johtanut sinut ongelmiin.
        {\footnotesize'This is not the first time that your mouth has lead you to problems.'}
        Tämä ei ole ensimmäinen kerta kun suusi on saanut sinut vaikeuksiin,
        vai onko?                                                            
        {\footnotesize'This is not the first time that your mouth has lead yo to trouble, or is it?'} →4s
    \end{alltt}

\item Paraphrase pairs with habitual and recurring events: Always doing something is considered more precise than usually doing or having a custom of doing, but if one always does something he/she also has a custom of doing it, thus the subsumption flag is used.

    \begin{alltt}
        Kysyn aina silti. Se on sellainen tapa.
        {\footnotesize'I always ask anyway. It's a kind of habit.'}
        Mutta minulla on tapana kysyä sitä.    
        {\footnotesize'I have the custom of asking about it.'} →4> 

        Teen niin aina 
        {\footnotesize'I do like that always'}
        Teen yleensä niin 
        {\footnotesize'I usually do that.'} →4>
    \end{alltt}

\item Relative time vs. absolute time: If relative time is compared against absolute time which can be considered “world knowledge”, the absolute time is considered more general, however, when comparing two different relative times or relative time against absolute time which cannot be considered world knowledge, the label is 3.

    \begin{alltt}
        Matti Nykänen kuoli viime vuoden helmikuussa 55-vuotiaana.
        {\footnotesize'Matti Nykänen died at the age of 55 February last year.'}
        Matti Nykänen kuoli helmikuussa 2019. Hän oli kuollessaan 55-vuotias. 
        {\footnotesize'Matti Nykänen died February 2019. He was 55 years old when he passed away.'} →4>

        \emph{Compare to}
        Rikos tapahtui viime vuoden helmikuussa.
        {\footnotesize'The crime took place in February last year.'}
        Rikos sattui helmikuussa 2019.           
        {\footnotesize'The crime took place in February 2019.'} →3

        \emph{Compare to} 
        Moroo, lähren Thaimaahan reppuni kanssa kolmen viikon päästä.
        {\footnotesize'Hi, I'll be leaving for Thailand with my backpack in three weeks from now.'} 
        Morjes,lahden Thaimaahan itseni ja reppuni kanssa joulukuun alussa. 
        {\footnotesize'Hello, I'll be taking myself and my backpack to Thailand at the beginning 
          of December.'} →3
    \end{alltt}
    
\item The lack of time related titles in the other utterance typically causes the subsumption flag.

    \begin{alltt}
        Presidentti Niinistö matkusti
        {\footnotesize'President Niinistö traveled'}
        Niinistö matkusti             
        {\footnotesize'Niinistö traveled'} →4>

        \emph{Compare to}
        Sussexin herttua Harry
        {\footnotesize'Harry, Duke of Sussex'}
        Britannian Harry      
        {\footnotesize'Harry of Britain'} →4 
    \end{alltt}

\item The subsumption flag is used in typos when there clearly is a word/character missing from an utterance, and there is ambiguity as to which word/character it is, i.e. the meaning of the utterance would change depending on which word/character is put in place of the missing one.

    \begin{alltt}
        Tahdot uskoa että hän oli poliisi eikä joku hullu joka oli
        pukeutunut niin.
        {\footnotesize'You want to believe that he/she was a police officer and not some crazy person
          dressed up as one.'}
        Halua_ uskoa, että hän oli poliisi, eikä poliisiksi pukeutunut hullu. 
        {\footnotesize'Wanting to believe that he/she was a police officer and not some crazy person
          dressed up as one.'} →4>

        \emph{Compare to}
        Sinä et tiedä mitään
        {\footnotesize'You don't know anything'}
        Sinä tiedä mitään   
        {\footnotesize'You know nothing.'} →4
    \end{alltt}
    
\end{itemize}

\section{Style flag (s)}

\begin{itemize}

\item The style flag is used when there is a notable difference in text registers e.g. written standard language, spoken language, slang or swear words, which may limit the usage of the utterance in certain contexts.
\item The style flag is used also when hedging, uncertainty or different levels of politeness or tone are present.
\item A general rule of thumb for distinguishing if a difference is markable is to imagine, when given one utterance, would the machine generating the other be desirable.

    \begin{alltt}
        Helou gimmat
        {\footnotesize'Hey U all girlz'}
        Päivää tytöt 
        {\footnotesize'How do you do girls'} →4s
    \end{alltt}
    
\item The other utterance uses pejorative or spoken language, while the other does not.

    \begin{alltt}
        Hän pahoinpiteli neljä poliisia sairaalakuntoon.
        {\footnotesize'He/she assaulted four police officers in such a way that they needed
          hospitalisation.'}
        Hän hakkasi neljä kyttää sairaalakuntoon.       
        {\footnotesize'He/she assaulted four coppers in such a way that they needed hospitalisation.'}→4s

        Mistä minulla olisi 100 000 dollaria?
        {\footnotesize'How could I have 100 000 dollars?'}
        Mistä helvetistä löydän 100 000 dollaria? 
        {\footnotesize'Where the hell will I find 100 000 dollers?'} →4s

        \emph{Compare to}
        Hiljaa, paskiaiset!
        {\footnotesize'Quiet, you bastards!'}
        Turpa kiinni, kusipäät! 
        {\footnotesize'Shut up you assholes'} →4

        \emph{Compare to}
        Vitustako minä tiedän.
        {\footnotesize'How the fuck should I know.'}
        Mistä pirusta minä tiedän. 
        {\footnotesize'How the hell should I know.'} →4
    \end{alltt}
    
\item There is a noticeable difference in strength or intensity of the utterances (tone), while the semantic meaning and the aspect strengthened stays equivalent.

    \begin{alltt}
        Hän on vanha.
        {\footnotesize'He/she is old.'}
        Hän on todella vanha. 
        {\footnotesize'He/she is extremely old.'} →4s

        Mitä katsot? 
        {\footnotesize'What are you looking at?'}
        Mitä tuijotat? 
        {\footnotesize'What are you gawking at?'} →4s

        Täällä on kylmä ilmapiiri
        {\footnotesize'The atmosphere here is cold'}
        Täällä on jäätävä tunnelma 
        {\footnotesize'The ambiance here is frosty'} →4s

        \emph{Compare to}
        Hän tuijottaa sinua vihaisesti
        {\footnotesize'He/she is staring at you hatefully'}
        Hän mulkoilee sinua vihaisesti 
        {\footnotesize'He/she is gawking at you hatefully'} →4
    \end{alltt}

\item Use the style flag also when there is a noticeable difference in politeness, many times marked with conditional verb mood.

    \begin{alltt}
        Arlene Fowler, soisitko sen kunnian, että tulisit vaimokseni?
        {\footnotesize'Arlene Fowler, will you give me the honour of becoming my wife?'}
        Arlene Fowler, menetkö kanssani naimisiin?                   
        {\footnotesize'Arlene Fowler, will you marry me?'} →4s

        Valitse sinä puheenaihe.
        {\footnotesize'You choose the topic to talk about.'}
        Entä jos sinä valitsisit puheenaiheen? 
        {\footnotesize'How about you choosing the topic to talk about?'} →4s
    \end{alltt}

\item Uncertainty and hedging is marked with the style flag if the other utterance is a confident statement. 

    \begin{alltt}
        Se taitaa tarvita muutenkin lisää tilaa.
        {\footnotesize'It might need more space anyway.'}
        Se tarvitsee muutenkin lisää tilaa.     
        {\footnotesize'It will need more space anyway.'} →4s

        en ehkä tiedä
        {\footnotesize'perhaps I don't know'}
        en tiedä      
        {\footnotesize'I don't know'} →4s

        Pakettini saapuu tänään.
        {\footnotesize'My package will arrive today.'}
        Pakettini saapuu varmaan tänään. 
        {\footnotesize'My package till probably arrive today.'} →4s
    \end{alltt}

\item Tag questions are primarily marked with the style flag (uncertainty), but if the other utterance has a genuine question whereas the other being certain or uncertain statement, the subsumption flag is used instead.

    \begin{alltt}
        Et nähnyt sitä.
        {\footnotesize'You did not see it'}
        Et nähnyt sitä, ethän?  
        {\footnotesize'I hope you did not see that?'} →4<

        \emph{Compare to}
        Etkö sinä nähnyt sitä?
        {\footnotesize'You didn't see that?'}
        Et nähnyt sitä, ethän?  
        {\footnotesize'I hope you did not see that?'} →4<
    \end{alltt}

\end{itemize}

\section{Minor deviation flag (i)}

\begin{itemize}

\item Small differences in certain grammatical categories are marked with the minor deviation flag if they change the semantic meaning of the utterance, however the meaning of the whole utterance staying essentially the same (e.g. this vs. that).
\item Such categories include e.g. tense, mood, person and grammatical number.
\item The minor deviation flag is used also in cases where both utterances have a slightly different place adverb, meaning being essentially the same.
\item However, if the difference is purely grammatical and does not change the meaning, the minor deviation flag should not be marked. These cases can include e.g. different past tenses, or idioms and phrases where changing the number of the verb would create ungrammatical sentences.

    \begin{alltt}
        Tule tänne
        {\footnotesize'You come here'}
        Tulkaa tänne 
        {\footnotesize'You guys come over here'} →4i 
        
        Hän on täällä.
        {\footnotesize'He is here.'}
        Hän on tuossa 
        {\footnotesize'He is there.'} →4i

        Parkkeerasimme tähän
        {\footnotesize'We parked here'}
        Parkkeeraamme tähän 
        {\footnotesize'We will be parking here'} →4i

        \emph{Compare to}
        Minulla on ajatus
        {\footnotesize'I have got a thought'}       
        Sain idean        
        {\footnotesize'I have an idea'} →4
        

        \emph{Compare to}
        Kaikki kuolevat ennen pitkää.
        {\footnotesize'Eventually everyone will die.'}
        Jokaisen on kuoltava joskus. 
        {\footnotesize'Everybody has to die at some point.'} →4
    \end{alltt}

\item The difference between "sinä" and "te" is generally marked with the minor deviation flag (not the style flag for politeness) if the difference between formal addressing and plural is not clear. However, if the difference between second person plural and formal addressing is obvious, the style flag should be used instead.
    
    \begin{alltt}
        Mihin te piiloutuisitte
        {\footnotesize'Where would you guys hide?'}
        niin minne piiloutuisit? 
        {\footnotesize'Where will you hide?'} →4i

        \emph{Compare to}
        Olet maineikas mies.
        {\footnotesize'You are a remarkable man, Mr.'}
        Olette maineikas mies. 
        {\footnotesize'You, sir, are a remarkable man.'} →4s

        \emph{Compare to}
        Tulin itse asiassa tapaamaan vaimoasi.
        {\footnotesize'I actually came here to meet your wife.'}
        Tulin oikeastaan tapaamaan vaimoanne. 
        {\footnotesize'I actually came here to meet your wife, Sir.'} →4s
    \end{alltt}
    
\item The minor deviation flag is added when both of the utterances have a different pronoun/possessive suffix. An exception is made with utterances comparing "hän" and "se", where "se" can refer to several entities (person, animal, thing), while "hän" can only be a person reference causing the subsumption flag instead. 

    \begin{alltt}
        Tämä laite on epäkunnossa.
        {\footnotesize'This apparatus is out of order.'}
        Tuo kone on rikki.        
        {\footnotesize'That apparatus is broken.'} →4i

        \emph{Compare to}
        Eiväthän he ammu häntä, eihän?
        {\footnotesize'They will not shoot him, I hope so?'}
        Ei kai ne ammu sitä?
        {\footnotesize'surely, it will not be shot at by those?'} →4>
    \end{alltt}

\item One utterance makes a clear reference to future time, while the other is in present or past time.

    \begin{alltt}
        Minä kävelen kotiin
        {\footnotesize'I'm walking home'}
        Minä aion kävellä kotiin 
        {\footnotesize'I plan to walk home'} →4i
    
        Me teemme töitä 
        {\footnotesize'We are working'}
        Me tulemme tekemään töitä 
        {\footnotesize'We came to work'} →4i 

        Miksi Hammond sanoi, ettei muista Tyleriä?
        {\footnotesize'Why did Hammond way that he/she doesn't remember Tyler?'}
        Miksi Hammond sanoisi ettei muista Tyleriä? 
        {\footnotesize'Why would Hammond be stating not recalling Tyler?'} →4i

        Meidän piti alkaa pitää huolta toisistamme.
        {\footnotesize'We were supposed to start looking after each other.'}
        Meidän täytyisi alkaa katsoa toistemme perään. 
        {\footnotesize'We should start looking after each other.'} →4i

        \emph{Compare to}
        Minä tapan sinut.
        {\footnotesize'I am killing you.'}
        Minä aion surmata sinut. 
        {\footnotesize'I plan on slaughtering you.'} →4
    \end{alltt}  
    
\item Within the past tenses, the minor deviation flag is only marked when there is a clear difference in meaning.

    \begin{alltt}
        Hän oli presidenttinä kuusi vuotta.
        {\footnotesize'He/she served as a president for six years.'}
        Hän on ollut presidenttinä jo kuusi vuotta. 
        {\footnotesize'He/she has been president for six years.'} →4i
        
        \emph{Compare to}
        Ei sinua unohdettu tänne.
        {\footnotesize'You have not been forgotten here.'}
        Ei sinua ole unohdettu tänne. 
        {\footnotesize'You are not forgotten here'} →4
    \end{alltt}

\end{itemize}

\section{Label 3: Context dependent paraphrases}
\begin{itemize}

\item The utterances are paraphrases in the given context, but not in all non-absurd situations.
\item Includes e.g. utterances where both have different ambiguous meaning or different additional information present.
\item When there is a conflict in the direction of the subsumption flag, assign label 3.

    \begin{alltt}
        Miten eilisilta meni?
        {\footnotesize'How did it go last evening?'}
        Miten teillä meni eilen? 
        {\footnotesize'How did it go yesterday?'} →3

        Milloin sait tietokoneen?
        {\footnotesize'When did you get the computer?'}
        Koska hankit koneen?     
        {\footnotesize'When did you acquire a machine?'} →3

        911. 
        Hätänumero?
        {\footnotesize'The emergency services number?'} →3
        
        Mitä jos keulisimme vähän?
        {\footnotesize'How about if we do a wheelie?'}
        Kärytetäänkö vähän kumia? 
        {\footnotesize'Should we burn some rubber?'} →3
    \end{alltt}

\item Uncustomary metaphors that require context to be understood. Can be direct translations so they need context from the original language and culture. The utterance must still be understandable in some way, if the utterance is not understandable in Finnish due to e.g. literal translation, label 2 is assigned.

    \begin{alltt}   
        Hän on kuin pieni jäätelöpallo.
        {\footnotesize'He/she is like a ball of ice cream.'}
        Hän on suorastaan syötävän suloinen. 
        {\footnotesize'He/she is frankly so adorable I could just eat him/her up.'} →3

        kaksi kärpästä yhdellä iskulla 
        {\footnotesize'I'll get two things done at the same time'}
        kaksi lintua yhdellä kivellä. 
        {\footnotesize'two birds with one stone'} →3
        
        \emph{Compare to}
        Olet löytänyt onnen
        {\footnotesize'You have found happiness'}
        Nyt sinulla on avaimet linnaan 
        {\footnotesize'Now you have the keys to the castle'} →2
    \end{alltt}

\end{itemize}

\section{Label 2: Related, but not a paraphrase}

\begin{itemize}

\item The utterances are related, but could hardly be paraphrases in non-absurd situations.
\item Utterances are discussing the same event/matter or they are topically similar, but they convey different information.
\item There is a significant difference in the main message even if similar additional information is present.

    \begin{alltt}
        Hallitusohjelma on saatu valmiiksi, julkistetaan maanantaina
        {\footnotesize'The government programme has been completed, it is due to be published on Monday'}
        Puheenjohtajat palaavat neuvottelupöytään Säätytalolle       
        {\footnotesize'The leaders will return to he negotiation table at House of the Estates'} →2

        Kersantti?
        {\footnotesize'Sergeant?'}
        Ylikonstaapeli. 
        {\footnotesize'Police sergeant.'} →2
    \end{alltt}

\item Contradictory information in utterances. The utterances can be in a different chronological viewpoint. The situation has changed substantially between the utterances.

    \begin{alltt}
        Merestä löytyneet lentokoneen osat varmistuivat kadonneen chileläiskoneen
        jäänteiksi
        {\footnotesize'The airplane parts found in the ocean have been confirmed to be from the wreck
          of a plane from Chile'}
        Kadonneeseen chileläiskoneeseen viittavia jäänteitä löydettiin merestä  
        {\footnotesize'Some place wreck material has been found in the ocean that could hint at the
          plane from Chile'} →2

        Useita loukkaantunut tulivuorenpurkauksessa Uudessa-Seelannissa
        {\footnotesize'Several have been injured at a volcano eruption in New Zealand'}
        Ei elonmerkkejä – Uuden-Seelannin White Islandin tulivuorenpurkauksessa
        on kuollut ainakin viisi ihmistä ja kateissa on useita                  
        {\footnotesize'No sings of life - At least five people have been killed and several missing
          after a volcano eruption on White Island in New Zealand'} →2 

        \emph{Compare to}
        Viisi kuollut ja useita kateissa tulivuorenpurkauksen jäljiltä
        {\footnotesize'Five people have been killed and several missing after volcano eruption'}
        Useita uhreja tulivuorenpurkauksen jäljiltä                    
        {\footnotesize'Several victims after a volcano eruption'} →4>
    \end{alltt}
    
\item Utterances can be paraphrases in a certain context but not in the given situation.

    \begin{alltt}
        Selaa eteenpäin. [Tämä nettisivu sisältää tarvittavan tiedon.]
        {\footnotesize'Scroll down. [This website will contain the information needed.]'}
        Rullaa alas. [Tämä rinne on jyrkkä.]
        {\footnotesize'Roll down. [This is a steep hill.]'}  →2
    \end{alltt}

\end{itemize}

\section{Label 1: Unrelated}

\begin{itemize}

\item There is no apparent link between the two utterances.
\item Usually can be considered as candidate selection errors.
\item If the candidate pair shares only a single proper name while the topic otherwise is different, the candidate is considered unrelated.

    \begin{alltt}
        Oletteko Sherlock Holmes?
        {\footnotesize'Are you Sherlock Holmes?'}
        Riippuu.
        {\footnotesize'That depends.'} →1

        Sipoonranta on Sipoossa, ei Helsingissä
        {\footnotesize'Sipoonranta is located in Sipoo not Helsinki'}
        Sipoonranta hakee taas lisäaikaa rakentamiseen 
        {\footnotesize'Sipoonranta is once more requesting additional time for the build'} →1
    \end{alltt}

\end{itemize}

\section{Label x: Skip}

\begin{itemize}

\item Serious problems in the data, i.e. wrong language.
\item Although seriously considered cannot decide a label, e.g. if the utterance is not understandable.
\item Assigning a label would not serve the desired purpose, e.g. the utterances are identical.
\item Should be used sparingly.

\end{itemize}

\section{Additional guidelines}

\subsection{Rules for changing the original text in automatically extracted paraphrase candidates}

A small fraction of the Turku Paraphrase Corpus is constructed by manually annotating automatically extracted paraphrase candidates originating from news headings. As news headings are often constructed from two or more semantically independent elements separated clearly with punctuation characters such as semicolon or hyphen, in order to make a paraphrase it is often desired to select only some of such independent elements. However, in the annotation workflow of the corpus it would be inefficient to run the headings through the manual paraphrase extraction, thus these automatically extracted candidates are presented to the annotators only during the paraphrase labeling phase. In these cases the annotators are allowed to change the original text in the paraphrase annotation tool. However, strict rules are applied.

\begin{itemize}

\item Delete only semantically independent parts that are separated from the rest of the sentence with some punctuation character(s) such as semicolon or hyphen.
    
\item If there is something independent, which is clearly separated with punctuation characters in the middle of the sentence, the part can be deleted as well. This can happen mostly in cases, where the heading is constructed of three independent elements.

\item Do not delete the element if it is not necessary, however, independent elements which do not have any correspondence in the paraphrase pair should be deleted in order to avoid overuse of the subsumption flag.

\item Do not delete single words that are in the middle of the sentence, you can do this in the rewrite section. For details, see Section~\ref{sec:rew} below.

\item Do not delete coordinate or subordinate clauses, you can do this in the rewrite section.

\item Do not delete the element if the remaining sentence becomes ungrammatical or clumsy, this means the element is not independent.

\item Do not add anything in the original utterances, you can do this in the rewrite section.

\item If the changes to the original utterances create an identical paraphrase pair (the utterances become identical), assign x (skip) and rewrite the pair if possible for example by changing the order of the words.

    \begin{alltt}
        Irakin levottomuudet jatkuvat – AFP: Shiiajohtajan kotia pommitettiin
        lennokista 
        {\footnotesize'The unrest is continuing in Iraq - AFP: The Shia leader's home was bombed from a
          miniature aircraft'}
          
        → Irakin levottomuudet jatkuvat: Shiiajohtajan kotia pommitettiin
        lennokista
        {\footnotesize'The unrest is continuing in Iraq: The Shia leader's home was bombed from a
          miniature aircraft'}

    \end{alltt}

\end{itemize}

\subsection{Rules for rewriting}
\label{sec:rew}

The paraphrase annotation tool supports for rewriting the examples in addition to annotating the label for the original candidate. Note that rewriting keeps the original paraphrase pair unchanged, and creates an additional paraphrase pair. Annotators are instructed to rewrite paraphrase pairs in order to turn the candidates into universal paraphrases in cases where the original label is something else than label 4 without any flags. Rewriting is suitable for candidates where a simple edit, for example word or phrase deletion, addition or re-placement with a synonym or changing an inflection, can be used to create an universal paraphrase. However, if the rewriting would require more complicated changes, and a good solution does not come into mind easily, it's preferred to skip the rewrite option in order to use the time efficiently. Rewrites must be such that the annotated label for the rewritten example is always label 4. In addition to rewriting, the original candidate is labeled as usually.

Rewriting process is quite loosely defined, however, following guidelines are used to guide the process.

\begin{itemize}

\item Rewrite the paraphrase candidates if it is relatively simple to do so and does not take too much time.

\item Make always as few changes as possible, if changing a word is not necessary in order to make a universal paraphrase or create more variation to the paraphrase pair, do not change it.

\item Try to avoid repeating the same terms as in the other utterance, prefer using e.g. synonyms.

\item Deleting additional information is preferred if the sentence stays natural and it does not remove all the variation from the paraphrase pair.

\item Inserting additional information is also possible if deleting it from the other utterance does not feel suitable solution, however, try not to plainly copy the words from the other sentence. If the utterances get more varied from adding the missing information, it is allowed to do so.

\item Do not unnecessarily change the meaning of the original utterances.

\item Try to avoid using a systematic style.

\item Do not fix the typos during rewriting if it is not necessary for the meaning (remove ambiguity).

\item During rewriting, it's preferred to keep the original style if possible, try not to make unnecessary edits like correcting capitalization or punctuation.

\end{itemize}

\subsection{General guidelines for paraphrase extraction}

Most of the paraphrases in the Turku Paraphrase Corpus are manually extracted from related, paraphrase heavy source documents (such as alternative translations of the same source texts, news articles reporting on the same event, or related discussion forum messages seeking information from the same topic or sharing related experiences). The manual paraphrase extraction process is carried out with relatively loose instructions. However, general guidelines are provided in order to avoid extracting a large amount of trivial paraphrase candidates including only element variation, such as simple word reordering or synonym replacement.

\begin{itemize}

\item Extract as long continuous utterances as possible. The extracted paraphrase can be anything between a short phrase or long text span including several sentences. However, do not over-extend the candidate by selecting content not present in the another utterance.

\item Avoid extracting a large amount of trivial paraphrase pairs with only elementary variation, such as differing only by word order, one synonym or inflectional ending, if not otherwise considered especially interesting. 

\item Do not actively extract pairs where the only difference is singular or plural form, personal pronoun or punctuation.

\item Even if common phrases, such as "Hei" or "Kiitos" would create a good paraphrase, do not extract such if you clearly remember already extracting the same pair previously.

\item If pairs with only elementary variation are accidentally extracted, they can be annotated with the label x in the paraphrase annotation phase. 

\item The following pairs are examples of elementary variation not actively extracted:

    \begin{alltt}
    
        Päästäkää minut!
        {\footnotesize'Let me go!'}
        Päästä minut!
        {\footnotesize'Let go of me!'}

        Ei teillä ole todisteita.
        {\footnotesize'You folks don't have evidence.'}
        Sinulla ei ole todisteita.
        {\footnotesize'You don't have evidence.'}
        
        Kuka minä sitä olen muuttamaan.
        {\footnotesize'Who am I to change that.'}
        Kuka minä olen sitä muuttamaan?
        {\footnotesize'Who am I to change that?'}
        
        Katson 
        {\footnotesize''m looking'}
        Minä katson
        {\footnotesize'I am looking'}
    \end{alltt}
    
\end{itemize}

\section*{Acknowledgments}

The Turku paraphrase corpus project was supported by the European Language Grid project through its open call for pilot projects. The European Language Grid project has received funding from the European Union’s Horizon 2020 Research and Innovation programme under Grant Agreement no. 825627 (ELG). Computational resources were provided by CSC — the Finnish IT Center for Science and the research was supported by the Academy of Finland. We also thank Sampo Pyysalo for fruitful discussions and feedback throughout the project.

\bibliographystyle{unsrtnat}
\bibliography{references}  

\end{document}